\documentclass[10pt,twocolumn,letterpaper]{article}

\usepackage[pagenumbers]{cvpr} 

\usepackage[dvipsnames]{xcolor}

\definecolor{cvprblue}{rgb}{0.21,0.49,0.74}
\usepackage[pagebackref,breaklinks,colorlinks,citecolor=cvprblue]{hyperref}

\usepackage{times}
\usepackage{graphicx}
\usepackage{epstopdf}
\usepackage{amsmath}
\usepackage{amssymb}
\usepackage{multirow}
\usepackage{svg}
\usepackage{cuted}
\usepackage{caption}
\usepackage{float}
\usepackage{booktabs}
\usepackage{bbding}
\usepackage{color}
\newcommand{\red}[1]{\textcolor{red}{#1}}

\newcommand{\blue}[1]{\textcolor{blue}{#1}}
\definecolor{linkcolor}{RGB}{255,0,0}
\definecolor{urlcolor}{RGB}{255,105,180}
\definecolor{citecolor}{RGB}{66,168,235}
\newcommand \footnoteONLYtext[1]
{
	\let \mybackup \thefootnote
	\let \thefootnote \relax
	\footnotetext{#1}
	\let \thefootnote \mybackup
	\let \mybackup \imareallyundefinedcommand
}
\usepackage[misc]{ifsym}

\usepackage{ulem}

\def\paperID{113} 
\def\confName{3DV\xspace}
\def\confYear{2024\xspace}

\title{ModelNet-O: A Large-Scale Synthetic Dataset for Occlusion-Aware Point Cloud Classification}

\vspace{-1em}
\author{Zhongbin Fang$^{1}$, Xia Li$^{2}$, Xiangtai Li$^{3}$, Shen Zhao$^{1,~\textrm{\Letter}}$, Mengyuan Liu$^{4,~\textrm{\Letter}}$ \\
 \small{$^{1}$Sun Yat-sen University}
  \small{$^{2}$Department of Computer Science, ETH Zurich}
  \small{$^{3}$S-Lab, Nanyang Technological University} \\
  \small{$^{4}$Key Laboratory of Machine Perception, Shenzhen Graduate School, Peking University}
}

\begin{document}

\maketitle

\begin{strip}
\begin{figure}[H]
\hsize=\textwidth
\centering
\vspace{-6em}
\includegraphics[width=0.95\textwidth]{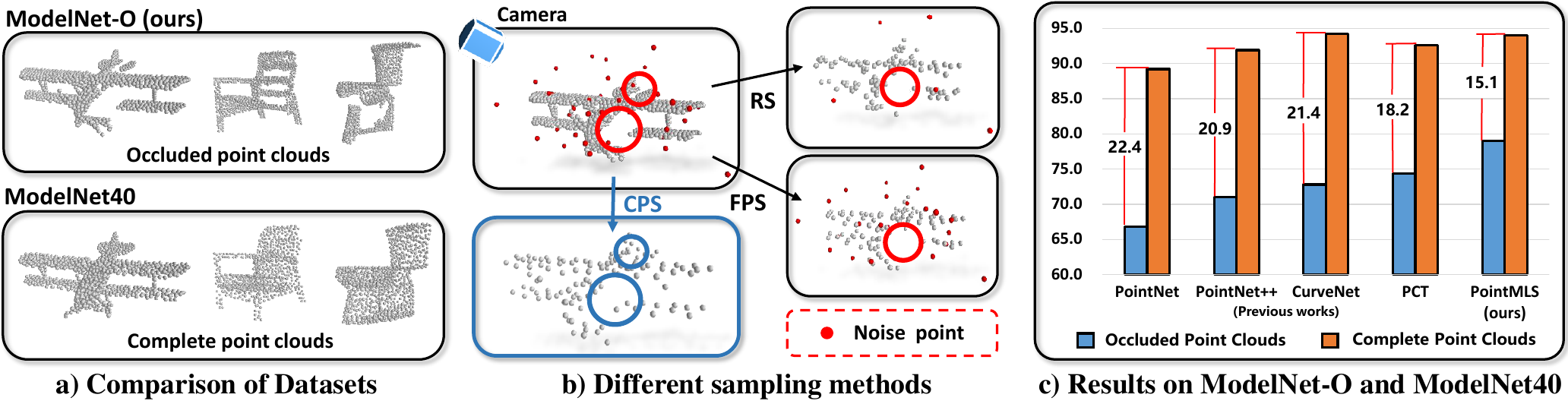}
\vspace{-1.5mm}
\caption{
a) Comparison between occluded point clouds (ModelNet-O) and completed point clouds~\cite{wu2015ModelNet40}, our ModelNet-O simulates the collection of point clouds using a fixed camera.
b) In the case of point sampling, previous methods~\cite{hu2020RandLa-Net,qi2017pointnet} suffer from instability in dealing with outlier points, while the proposed CPS module is more robust to noise and can weaken the effect of occlusion. c) Our proposed PointMLS performs well on both occluded~(ModelNet-O) and general~(ModelNet40) point clouds.}
\label{fig1}
\end{figure}
\vspace{-2em}
\end{strip}

\begin{abstract}
\vspace{-1em}
\footnoteONLYtext{\textrm{\Letter}~The corresponding authors are Shen Zhao and Mengyuan Liu.}Recently, 3D point cloud classification has made significant progress with the help of many datasets. However, these datasets do not reflect the incomplete nature of real-world point clouds caused by \textbf{\textit{occlusion}}, which limits the practical application of current methods.
To bridge this gap, we propose ModelNet-O, a large-scale synthetic dataset of 123,041 samples that emulate real-world point clouds with self-occlusion caused by scanning from monocular cameras.
ModelNet-O is \textbf{\textit{10 times}} larger than existing datasets and offers more challenging cases to evaluate the robustness of existing methods.
Our observation on ModelNet-O reveals that \textbf{well-designed sparse structures can preserve structural information of point clouds under occlusion}, motivating us to propose a robust point cloud processing method that leverages a critical point sampling (CPS) strategy in a multi-level manner. We term our method PointMLS. 
Through extensive experiments, we demonstrate that our PointMLS achieves state-of-the-art results on ModelNet-O and competitive results on regular datasets, and it is robust and effective.
More experiments also demonstrate the robustness and effectiveness of PointMLS.
Code is available at \url{https://github.com/fanglaosi/PointMLS}
\end{abstract}
\vspace{-1.5em}
\section{Introduction}

With the advent of 3D sensors such as LiDAR and Kinect, 3D point clouds have gained increasing popularity~\cite{wu2019pointconv,lu2020deeplearningsurvey,qi2018frustum,zhang2019rotation}, with the rapid progress in point cloud datasets and point cloud classification models~\cite{qi2017pointnet,qi2017pointnet++,wang2019dynamic,liu2019rscnn,guo2021pct,wu2019pointconv,thomas2019kpconv,simonovsky2017ECC,klokov2017escape,liu2019densepoint,xu2020GS-Net,yang2020cn}.
Recently, with the progress of vision transformer~\cite{vaswani2017attention}, several works~\cite{yu2022point-bert,pang2022point-mae,zhang2022point-m2ae,dong2022act,liu2019point2sequence,xie2018attentional} propose transformer-based approaches, showing close to state-of-the-art accuracy on public point cloud classification datasets such as ModelNet40~\cite{wu2015ModelNet40} and ScanObjectNN~\cite{uy2019scanobjectnn}. 
However, in real-world scenarios, point cloud often suffers from local area incompleteness due to self-occlusion and may contain noise distributed around the object due to monocular scanning. Furthermore, the 3D point cloud classification has a seriously lacking data.
As a result, many current methods focus on clean and complete point clouds, failing to account for occluded point clouds, which are prevalent in the real world.
Therefore, many point cloud classification models~\cite{qi2017pointnet++,guo2021pct} exhibit poor performance when dealing with \textit{occluded point clouds}, as shown in the left of the Fig.~\ref{fig1}. 
In other words, previous point cloud classification models are not robust enough to point clouds with self-occlusion.

In order to measure the robustness of the model, ModelNet-C~\cite{ren2022ModelNet-C} summarizes several atomic corruptions of point clouds and proposes two indicators, corruption error (CE) and relative corruption error (RCE),
to measure the model's performance on these corruptions, which significantly contributes to the robustness measure of point clouds. 
However, when collecting point clouds in the real world, only the points facing the cameras are accepted. 
And due to the occlusion of the front surface, the points behind will be abandoned. 
This occlusion situation is different from the conventional datasets, such as ModelNet40 and ShapeNet, and cannot be composed of atomic corruptions proposed by~\cite{ren2022ModelNet-C}.

In order to better understand the under-occlusion problem of point clouds, we introduce the ModelNet-O dataset that contains 123,041 occluded point cloud samples. 
It simulates the real point clouds collected by the sensor in the real scene via camera projection. 
Unlike traditional point cloud classification datasets, ModelNet-O contains partially missing point cloud objects due to self-occlusion. 
Moreover, we define precise criteria for occluded point clouds classification evaluation to standardize the evaluation of all the reported results on this benchmark. 
The training and testing sets of the ModelNet-O have \textit{different} camera views, respectively.
In short, ModelNet-O aims to measure the real-world classification performance on occluded point clouds. 

Based on ModelNet-O, we observe that a well-designed sparse structure is a crucial point to solving the above problem, 
which preserves structural information of the point clouds with self-occlusion and neglects noisy points
as shown in Fig.~\ref{fig1}. With this prior, we argue an effective sampling method is essential for the robustness of the proposed dataset.
Though many previous works~\cite{qi2017pointnet++,yang2019PAT,yan2020pointasnl} have explored this direction, their methods neither suffer from strong outlier-sensitiveness nor self-occlusion.

To address these issues, we propose a robust point cloud classification framework PointMLS with a multi-level critical points sampling mechanism (Sec.~\ref{sec:method}).
PointMLS consists of two main modules: the critical point sampling (CPS) module and the feature aggregation (FA) module. The role of the former is to sample the input dense point cloud into a sparse point cloud while preserving the structural information of the original point clouds.
Unlike previous methods, CPS combines point-wise and global features of point clouds, making it more robust against occlusion. 
Besides, The FA module hierarchically aggregates the local features of the point clouds and captures the context dependencies of different layers. 
Intuitively, the sparser the point cloud, the less the model is affected by occlusion, and the more information will be lost. 
For supplementing information of sparse point clouds at different sampling levels, we propose the multi-level sampling (MLS) architecture, which combines the key structure from the sparser levels and details spatial information for denser levels.

With the above design, PointMLS achieves an overall accuracy of \textbf{78.9$\%$}, outperforming representative classic 3D point cloud methods and the recent state-of-the-art methods on MoldeNet40. 
Furthermore, PointMLS shows strong robustness against different noise inputs (Sec.~\ref{sec:main_results}).

Our contributions can be summarized as follows: 
\begin{itemize}
    \item We introduce a challenging occlusion point cloud classification dataset \textbf{ModelNet-O} that better reflects real-world scenarios and contains large-scale data.
    \item We propose a robust point cloud classification method, \textbf{PointMLS}, based on a multi-level sampling strategy.
    \item PointMLS achieves \textbf{state-of-the-art} overall accuracy on the occlusion point cloud dataset ModelNet-O and achieves \textbf{competitive} accuracy on the regular datasets, ModelNet40 and ScanObjectNN.
\end{itemize}

\section{Related Work}

\noindent
\textbf{3D Point Clouds Classification.}
In recent years, various deep neural networks have been proposed for the 3D point cloud classification task. 
In particular, PointNet~\cite{qi2017pointnet} and PointNet++~\cite{qi2017pointnet++} are pioneers of point-based methods in 3D point cloud analysis work, proposing a deep neural network that consumes point clouds directly.
Moreover, graph-based methods~\cite{wang2019dynamic,liu2019lpdnet,jiang2019hierarchical,zhao2019pointweb,zhang2019unsupervised} apply geometric topology to the point cloud analysis task via Edge-Conv operator~\cite{simonovsky2017ECC}. To search for local geometric features of point clouds, several works~\cite{wu2019pointconv,thomas2019kpconv,shen2018kcnet,li2018pointcnn,xu2021paconv,komarichev2019acnn} propose different convolution kernels.
Recently, there has been a growing interest in exploring the shape description of the point cloud~\cite{liu2019rscnn,xu2021GDANet,xiang2021curvenet,ran2022surface} and different pre-training strategies~\cite{fu2022posbert,wang2022p2p,zhang2022pointclip, zhu2022pointclipv2,chen2023pointgpt,jiang2022mae3d,liu2022masked, xie2020pointcontrast,zhang2021DepthContras}. Within these methods, PointMLP~\cite{ma2022pointmlp} directly applies a pure residual MLP network to the 3D point cloud analysis. However, all of these methods cannot be directly applied to the under-occlusion point cloud setting,  no matter directly transferring or training from scratch.

\noindent
\textbf{Sampling Strategies.}
The goal of point cloud sampling is to remove noisy points and reduce the computation cost of large-scale point cloud analysis tasks.
Various sampling strategies~\cite{qi2017pointnet,hermosilla2018PDS,hu2020RandLa-Net,zhang2022IA-SSD,rakotosaona2020PointCleanNet,xu2020GS-Net,dell2022arbitrary} have been proposed in recent years.
For instance, FPS~\cite{qi2017pointnet} was first proposed and widely used in point cloud grouping processing. However, FPS is task-independent and sensitive to outliers.
To overcome the issue mentioned above, several adaptive sampling methods have been proposed, such as S-Net~\cite{dovrat2019S-Net}, PAT~\cite{yang2019PAT}, CP-Net\cite{nezhadarya2020CP-Net}, SampleNet~\cite{lang2020SampleNet}, and PointASNL~\cite{yan2020pointasnl}. S-Net applies a deep network to transform dense point clouds into simplified point clouds in a data-driven manner. PAT~\cite{yang2019PAT} and CP-Net~\cite{yan2020pointasnl} project the point cloud into the high-dimension feature space, then sample point clouds via Gumbel-softmax and sorting operations, respectively.
SampleNet and PointASNL explore the local neighborhoods with center points sampling via FPS and dynamically mix the point in the local region in a specific way.
Unlike the methods mentioned above, our critical point sampling module combines point-wise and global features to assign weights to each point for selecting critical points in a soft manner, completely out of the FPS.

\noindent
\textbf{Robustness Analysis and Benchmarks.}
Improving the robustness of models is critical in 3D point cloud classification~\cite{xiao2021triangle}.
To effectively remove the noise, Dup-Net~\cite{zhou2019Dup-Net} defines a denoiser while MaskNet~\cite{sarode2020masknet} uses a noise-free template point cloud.
Another solution is data augmentation, which is proved in a series of works~\cite{chen2020pointmixup,li2020pointaugment,kim2021pointWOLF,lee2021RSMix}.
To evaluate the effectiveness and robustness of the above method,
Robust-PointSet~\cite{taghanaki2020robustpointset} evaluates the robustness of point cloud classifiers with different corruptions. In contrast, ModelNet-C~\cite{ren2022ModelNet-C} summarizes seven kinds of common corruptions and proposes a robustness test suite with specific metrics based on the performance of DGCNN, providing a unified evaluation standard for the robustness of point cloud classification. Different from that, our ModelNet-O dataset simulates problems such as self-occlusion and local missing when point clouds are projected by a fixed camera. 
Therefore, our proposed ModelNet-O dataset is more suited to the point clouds collected in the real world.

\section{The ModelNet-O Dataset}
\label{section3}

\begin{figure}[t]
    \centering
    \begin{minipage}[!t]{1\linewidth}
        \subfloat[]{
        \includegraphics[width=0.99\linewidth]{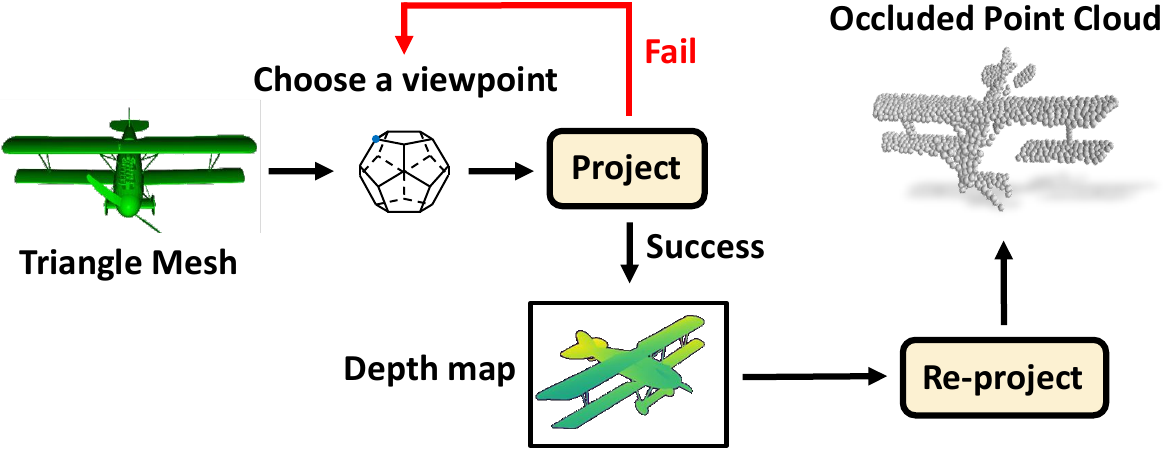}}
        \centering
    \end{minipage}
    \vspace{0.5em}
    
    \begin{minipage}[!t]{1\linewidth}
        \subfloat[]{ 
        \includegraphics[width=0.95\linewidth]{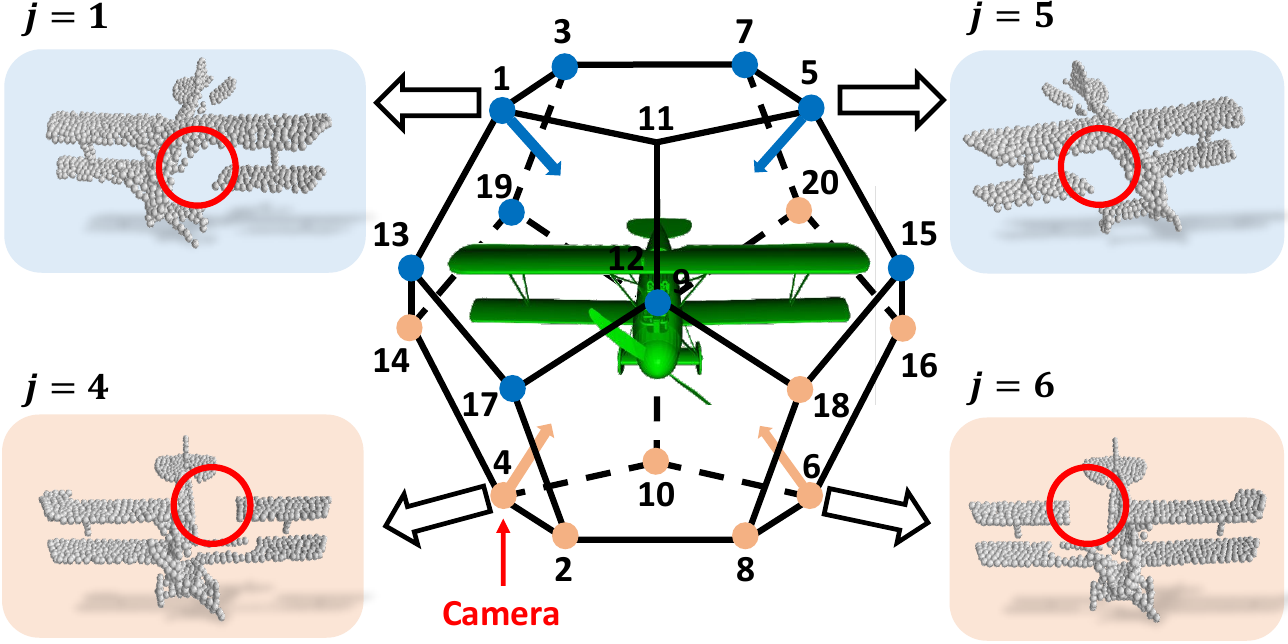}}
        \centering 
    \end{minipage}
    \vspace{-0.5em}
    \caption{a) The overall scheme of generating occluded point clouds. Note that some objects cannot be projected to obtain a depth map under certain viewpoints. b) 20-view dodecahedral configuration. The virtual cameras are placed on the vertices of a dodecahedron encompassing the object. Blue points: training set. Orange points: testing set.}
    \vspace{-0.5em}
    \label{fig3}
\vspace{-1em}
\end{figure}

Different from \cite{wu2015ModelNet40} and \cite{uy2019scanobjectnn}, the point clouds collected by 3D sensors in the real world are occluded, especially for real-time applications with monocular cameras. 
Thus, the collected points only cover the surface facing the 3D sensor, while leaving out the points behind the object. 
This creates a gap between current 3D point cloud classification tasks and real-world requirements.

To address this issue, we propose a challenging ModelNet-O dataset containing 123,041 occluded point cloud samples generated from  ModelNet40. Each sample in ModelNet-O is projected from one specific camera view, keeping the points facing the camera and discarding the behind ones. 
In this section, we introduce the generation process of occluded point clouds (Sec.~\ref{section3.1}) and define evaluation criteria for all reported results (Sec.~\ref{section3.2}).

\noindent
\subsection{Occlusion accomplishing}
\label{section3.1}

\begin{table}[t]
\caption{Comparison of ModelNet40, ScanObjectNN and ModelNet-O. ModelNet-O contains occluded point clouds and is almost 10$\times$ larger than ScanObjectNN.}
\vspace{-0.5em}
\centering
\small
\begin{tabular}{lccr}
\toprule[0.15em]
\textbf{Dataset}    & Occlusion  & Cross-View    & Total           \\ \hline
ModelNet40~\cite{wu2015ModelNet40}          & -           & -           & 12,311           \\
ScanObjectNN~\cite{uy2019scanobjectnn}        & \textbf{\checkmark}          & -           & 14,298           \\
\textbf{ModelNet-O~(ours)} & \textbf{\checkmark} & \textbf{\checkmark} & \textbf{123,041} \\ \bottomrule[0.1em]
\end{tabular}
\centering
\label{table1}
\vspace{-1.5em}
\end{table}

As shown in Fig.~\ref{fig3}(a), generating an occluded point cloud $P$ from a triangle mesh can be implemented by two consecutive steps: 1) Projecting the full point cloud through a fixed camera perspective to generate a depth map. 2) Reconstructing point cloud from the depth map.

\begin{figure*}[t]
\centering
\includegraphics[width=17cm]{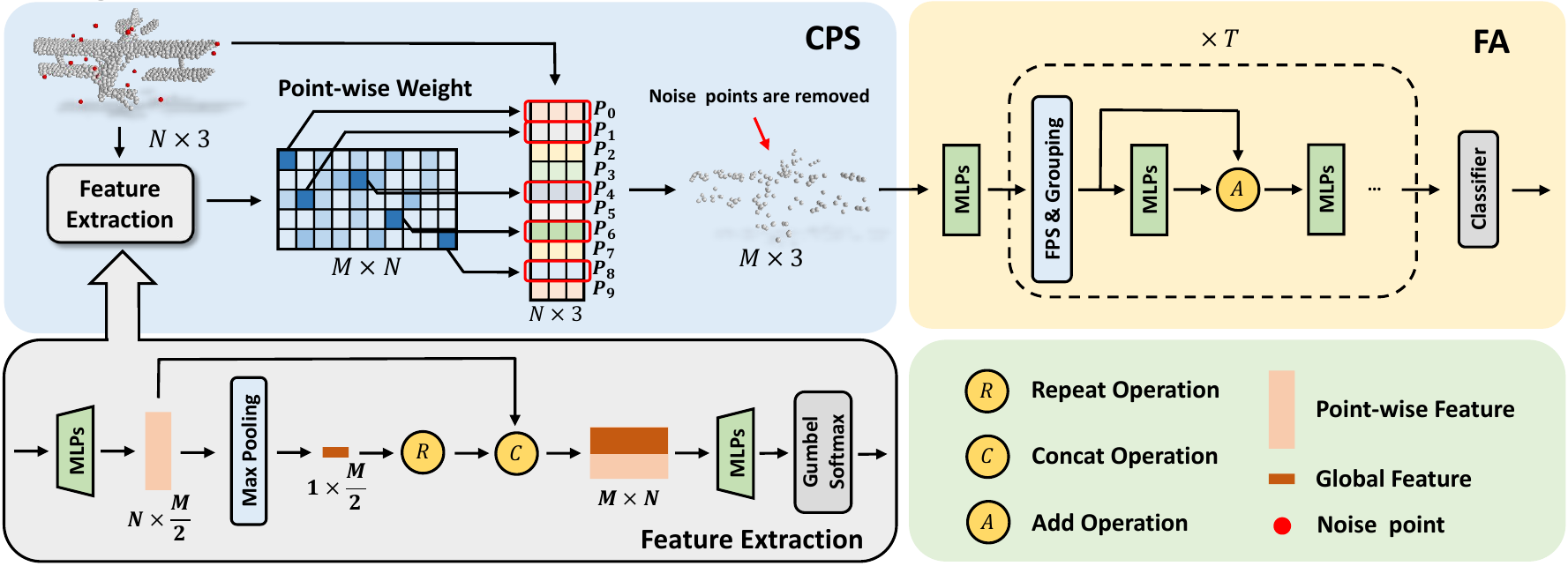}
\centering
\vspace{-0.5em}
\caption{Illustration of the critical point sampling (CPS) module and the feature aggregation (FA) module. The CPS module extracts features of the input point cloud $P\in \mathbb{R}^{N\times 3}$, and then obtains point-wise weight $W\in \mathbb{R}^{M\times N}$. Sampled point clouds $\tilde{P}\in \mathbb{R}^{M\times 3}$ are generated by matrix multiplication between $W$ and $P$. 
The FA module with residual MLPs further aggregates features of each local region belonging to the sampled point cloud. A classifier obtains the final score.
}
\vspace{-1.5em}
\label{fig2}
\end{figure*}

\noindent
\textbf{Projecting point cloud.} For illustration, we consider the simplest pinhole camera model. Based on which, the world coordinates $(x_{w},y_{w},z_{w})$ of a full point cloud  can be projected to the camera coordinate $(x_{c},y_{c},z_{c})$ via a simple linear transformation:
\begin{equation}
    \begin{bmatrix}
    x_{c}\\ 
    y_{c}\\ 
    z_{c}
    \end{bmatrix}=\begin{bmatrix}
    f_{x} &0  &u_{0}  &0 \\ 
    0 &f_{y}  &v_{0}  &0 \\ 
    0 &0  &1  &0 
    \end{bmatrix}\begin{bmatrix}
    \textbf{R} &\textbf{t} \\ 
    \textbf{0}^{T} &1 
    \end{bmatrix}\begin{bmatrix}
    x_{w}\\ 
    y_{w}\\ 
    z_{w}\\
    1
    \end{bmatrix},
    \label{eq1}
\end{equation}
where $\textbf{R}\in \mathbb{R}^{3\times 3}$ and $\textbf{t}\in \mathbb{R}^{3}$ are respectively rotation matrix and translation vector, which describe camera viewpoint. $\textbf{0}$ is a zero vector with size $\mathbb{R}^{3}$. What's more, parameters $(f_{x},f_{y})$ and $(u_{0},v_{0})$ in the camera intrinsic matrix denote the camera focal length and half the width and height of camera image.
Then, we may project the point with coordinate $(x_{c},y_{c},z_{c})$ to the location $(\frac{f_{x}x_{c}}{z_{c}},\frac{f_{y}y_{c}}{z_{c}})$ with depth value $z_{c}$  of the 2D depth map.

\noindent
\textbf{Generating occluded point cloud.}
After discarding the occluded points behind the surfaces, we can reconstruct the point cloud from 2D pixels of the depth map by the inverse transformation of Eq.~\ref{eq1}.
Such generated point cloud is a subset of the original one because of self-occlusion.

\subsection{Cross-View Evaluation}

\label{section3.2}
We take the 20-view dodecahedral configuration for the choice of viewpoints, as shown in Fig.~\ref{fig3}(b).
Then, we define the Cross-View evaluation criteria to standardize the evaluation of all reported results. Specifically, 20 camera viewpoints are split into two parts evenly, and we apply each to the training set $P_{Train}\left \{p_{j}|j=2i-1\right \}$ and testing set $P_{Test}\left \{p_{j}|j=2i\right \}$, 
where $i=1,2,...,10$ and $p_{j}$ denotes the occluded point cloud projected from the $j-th$ camera viewpoint. 
Since most viewpoints of the training set take the projections from the upside, while the testing set has never, the classification on ModelNet-O is challenging with the domain gap of camera views.

As Tab.~\ref{table1} displays, the ModelNet-O contains 123,041 occluded point clouds after removing unprojectable samples.
Unlike ModelNet40, ModelNet-O better simulates real-world scenarios that collect point clouds by a fixed camera.
ModelNet-O has almost \textbf{10 times} more samples than the ScanObjectNN dataset, which can help the community step forward in the 3D point cloud domain and makes it possible to apply data-hungry methods for this task.
\section{Our Approach: PointMLS}
\label{sec:method}
\noindent
\textbf{Overview.} To handle the problem induced by occlusion in point clouds, we introduce the model PointMLS, as presented in Fig.~\ref{fig2}.
It consists of three main components: critical point sampling (CPS) module (Sec.~\ref{Section4.1}), feature aggregation (FA) module (Sec.~\ref{Section4.2}), and multi-level sampling (MLS) architecture (Sec.~\ref{Section4.3}).

\subsection{Critical Point Sampling (CPS) Module}
\label{Section4.1}
For most existing methods, the farthest point sampling (FPS)~\cite{qi2017pointnet} is widely used due to its fast execution speed and appealing ability to extract the object contour. However, FPS is task-independent and outliers-sensitive, which requires a more robust sampling method.
At the same time, in our occlusion point cloud classification task, it would be promising if a sampling method could partially recover the occluded hole, keeping the structure of other parts while neglecting noisy points.
To handle these problems, we propose a Gumbel-softmax-based robust critical point sampling (CPS) module.

Given a specific input point cloud $\mathbf{P} \in \mathbb{R}^{N \times 3}$, as shown in Fig.~\ref{fig2}, we first encode them via Multi-Layer Perception layers (MLPs) to extract high-dimensional point-wise features and global features, which can be formulated as:
\begin{align}
    \mathbf{F}_{i}^{L}&=MLP_f(\mathbf{P}_{i})\in \mathbb{R}^{D}, \\
    \mathbf{f}^{G}&=Maxpooling(\mathbf{F}^{L})\in \mathbb{R}^{D},
\end{align}
\noindent
where $D$ is the output channel of MLPs, which is set to $\frac{M}{2}$. 
$\mathbf{F}_{i}^{L}$ represents a feature vector per point, while $\mathbf{f}^G \in \mathbb{R}^D$ indicates the global feature vector.

To make the sampling process differentiable and learnable, we adopt Gumbel-softmax~\cite{jang2016categorical} to assign a vector of sampling weights to each point:
\begin{equation}    \mathbf{W}_{i}=Gumbel\_softmax(MLP_w([\mathbf{F}_{i}^{L}\parallel \mathbf{f}^{G}]) / \tau),
\end{equation}
\noindent
where $\parallel $ is the concatenate operation and $\tau$ denotes the temperature parameter. 
We concatenate $\mathbf{F}_{i}^{L}$ and $\mathbf{f}^{G}$ and extract finer features through MLPs. 
During the sampling process, a specific point is generated according to each row of $\mathbf{W}$.
Finally, the final sample point cloud is generated by:
\begin{equation}
    \tilde{\mathbf{P}} = \mathbf{W} \mathbf{P}.
\end{equation}

Note that ${\left \{\tilde{\mathbf{P}}_{j}\right \}}_{j=0}^{M}\nsubseteq {\left \{\mathbf{P}_{i}\right \}}_{i=0}^{N}$. Since we use Gumbel-softmax for soft sampling, the sampled point cloud is not a subset of the input point cloud. This property allows the CPS module to recover the occluded region after sampling dense point clouds to sparse point clouds.

\subsection{Feature Aggregation (FA) Module}
\label{Section4.2}
To further classify the sampled point cloud $\tilde{P}$ after generating sampled point clouds, inspired by previous MLP-based network~\cite{ma2022pointmlp}, we propose a module simply combining MLPs with skip connections, termed the Feature Aggregation (FA) module. As shown in Fig.~\ref{fig2}, the sampled points ${\tilde{\mathbf{P}}}$ are first mapped to a high-dimensional feature space as $\tilde{\mathbf{F}}$,
followed by a K-NN operation to search for neighbors $\zeta (j)$ of the $j$-th sampled point in the original coordinate space. 
After that, each point feature is updated with the features of its neighborhoods using the following formulation:
\begin{equation}
    \tilde{\mathbf{F}}_{j}^{t+1}=Aggregation(\sigma (\tilde{\mathbf{F}}_{\zeta(j)}^{t})),t=0,1,...,T,
\end{equation}
where $\sigma(\mathbf{x}) = MLP(\mathbf{x})+\mathbf{x}$~\cite{ma2022pointmlp}. $\tilde{\mathbf{F}}_{\zeta(j)}$ is the collected features from neiborhoods of the $j$-th point. We implement the aggregation function as a Maxpooling operation. During the processing of the FA module, $\tilde{\mathbf{P}}$ and $\tilde{\mathbf{F}}$ are updated together for $T$ stages. The final prediction score $\mathbf{c}$ is generated by a classifier:
\begin{equation}
    \mathbf{c}=Classifier(\tilde{\mathbf{F}}^{T})
\end{equation}

\subsection{Multi-Level Architecture}

\label{Section4.3}
\begin{figure}[t]
\centering
\includegraphics[width=8.2cm]{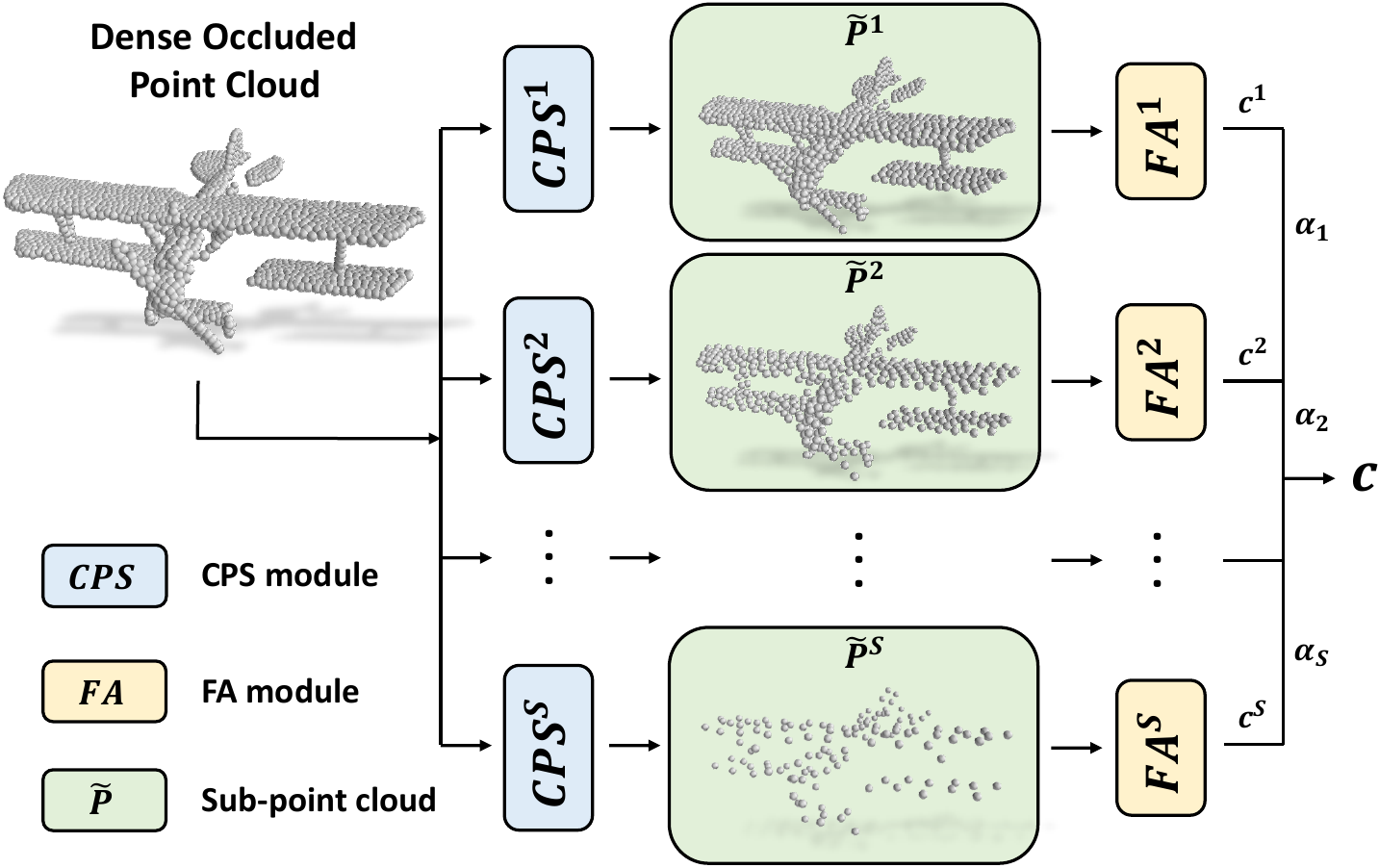}
\vspace{-0.5em}
\caption{Illustration of multi-level sampling (MLS).}
\vspace{-0.5em}
\label{fig4}
\end{figure}

With the proposed sampling strategy, we can obtain a tight representation of the points set. However, determining the optimal sampling ratio can be challenging as having more points provide more detailed spatial information, while sparser sets can reduce noise and compensate for the occlusion problem. Given this, diverse levels shall provide more modeling flexibility.
Thus, we propose a multi-level sampling (MLS) architecture. As shown in Fig.~\ref{fig4}, 
we sample the input point cloud with $S$ different ratios as a multi-level representation, each with the corresponding features. Then we feed each level into a separate FA module to obtain a prediction $\mathbf{c}^s$.
The final score $\mathbf{c}$ is formulated as:
\begin{equation}
    \mathbf{c} =\sum_{s=0}^{S}  \alpha_{s} \mathbf{c}^{s},
\end{equation}
where the composition ratio $\alpha_{s}$ is balance factor. 
We consider $S=4$ with levels of 1024, 512, 256, and 128 points for the experiments. 
Note that when the sampling level is smaller than 1024 points, the corresponding CPS module and FA module are designed to be lightweight.

\subsection{Training Loss}

As discussed above, PointMLS serves the sampling and classification tasks together. The overall training objective can be written as:
\begin{equation}
    L_{total}=L_{cls}(\mathbf{c},y)+L_{spl}(\mathbf{P},\tilde{\mathbf{P}}),
\end{equation}
where $y$ is the labels.
$L_{cls}$ is the cross-entropy loss of labels and predictions.
$L_{spl}$ is the Chamfer Distance of $\mathbf{P}$ and $\tilde{\mathbf{P}}$. Since our Gumbel-softmax outputs soft weights, the goal of $L_{spl}$ is to constrain the similarity between the initially generated point clouds and the input point clouds.
\section{Experiments}

\noindent
\textbf{Datasets and Metric.}
We evaluate our PointMLS on several different point cloud classification datasets, including both occluded point cloud datasets (ModelNet-O) and general point cloud datasets (ModelNet40 and ScanObjectNN). As discussed in Sec.~\ref{section3}, the ModelNet-O dataset comprises 98,369 training samples and 24,672 testing samples, which are assigned to 40 different categories. Similarly, the ModelNet40 dataset consists of 9,843 training and 2,468 testing complete point clouds, which are also classified into 40 categories. Unlike the ModelNet40 dataset, the point clouds in the ScanObjectNN have complex backgrounds. We perform experiments on the hardest perturbed variant (PB$\_$T50$\_$RS).

\noindent
\textbf{Implementation Details.}
Our multi-level architecture considers four sampling levels, with scales of 1024, 512, 256, and 128 points.
For extracting point-wise features of the original point cloud, we construct our CPS module using a multilayer perceptron ($MLP_{f}$) with dimensions [64, 128, 256, 512, 512].
We set $T=4$ stages in the FA module.
The temperature parameter $\tau$ of Gumbel-softmax is initialized to $1$ and gradually adjusted to $0.01$ using cosine annealing.
We set the initial learning rate to $0.1$ for ModelNet-O and ModelNet40, and $0.01$ for the ScanObjectNN.
The learning rate is dropped by a factor of $100$ during training.
We train all models for $65$ epochs.

\subsection{Main Results}
\label{sec:main_results}

\noindent
\textbf{Classification on Occluded Point Clouds.}
We evaluate the performance of recent representative methods on the occluded point cloud dataset ModelNet-O and compare the results with our proposed method, PointMLS.
For MAE-style methods~\cite{pang2022point-mae,dong2022act,zhang2023i2p-mae}, we use their pre-trained version and fine-tune the models on ModelNet-O.
As shown in Tab.~\ref{table2}, PointMLS achieves an overall accuracy of 78.9$\%$, outperforming the other methods.
Specifically, PointMLS outperforms PointNet~\cite{qi2017pointnet} by 13.2$\%$ in overall accuracy, while also surpassing the state-of-the-art method CurveNet~\cite{xiang2021curvenet} on ModelNet40 by 6.1$\%$.
Even without the multi-scale manner, our method also achieves 77.7$\%$ OA.

\begin{table}[t]
\caption{Class-average accuracy (mAcc) and overall accuracy (OA) on the proposed ModelNet-O dataset. $\eta~(\%)$ denotes the ratio of points replaced with noise. {[}P{]} denotes fine-tuning the models after self-supervised pre-training, which uses extra training data. $\dagger$ denotes the model trained without the Multi-level architecture. Bold: best results. Underline: second-best results.}
\vspace{-0.5em}
\centering
\small
\setlength\tabcolsep{0.7mm}
\begin{tabular}{l|c|cc|ccc}
\toprule[0.15em]
\multirow{2}{*}{Methods} & \multirow{2}{*}{Venues} & \multicolumn{2}{c|}{Clean}     & $\eta$=0.5   & $\eta$=2.5  & $\eta$=5    \\ 
                         &                             & mAcc   & OA      & OA      & OA        & OA       \\ \midrule
{[}P{]}PointMAE~\cite{pang2022point-mae} & ECCV’22     & 67.8   & 68.3    & 59.8    & 16.8      & 5.0      \\
{[}P{]}ACT~\cite{dong2022act}            & ICLR’23     & 69.2   & 69.7    & 61.8    & 28.1      & 8.2      \\
{[}P{]}I2P-MAE~\cite{zhang2023i2p-mae}   & CVPR’23     & 71.3   & 73.2    & 69.4    & 46.1      & 16.2     \\ \midrule
PointConv~\cite{wu2019pointconv}         & CVPR’19     & 65.0   & 65.7    & 64.5    & 48.4      & 26.7     \\
PointNet~\cite{qi2017pointnet}           & CVPR’17     & 65.6   & 66.8    & 47.0    & 20.0      & 10.5     \\
PointNet++~\cite{qi2017pointnet++}       & NeurIPS’17  & 70.7   & 71.0    & 67.6    & 44.6      & 20.8     \\
CurveNet~\cite{xiang2021curvenet}        & ICCV’22     & 71.9   & 72.8    & 65.8    & 30.9      & 8.4      \\
PCT~\cite{guo2021pct}                    & CVM’21      & 72.4   & 74.4    & 66.5    & 43.8      & 20.6     \\
DGCNN~\cite{wang2019dynamic}             & TOG’19      & 74.7   & 75.9    & 71.8    & 62.6      & 52.3     \\
PointMLP~\cite{ma2022pointmlp}           & ICLR’22     & 76.5   & 76.9    & 73.8    & 51.4      & 16.7     \\
PointMeta~\cite{lin2023pointmetabase}    & CVPR'23     & 76.4   & 77.0    & 72.5    & 52.7      & 22.8     \\
\midrule
\textbf{PointMLS$^{\dagger}$~(ours)} & - & \underline{76.7} & \underline{77.7} & \underline{77.5} & \underline{75.0} & \underline{68.5} \\
\textbf{PointMLS~(ours)}   & -     & \textbf{77.6} & \textbf{78.9} & \textbf{78.7} & \textbf{76.7} & \textbf{72.5} \\ \bottomrule[0.1em]
\end{tabular}
\centering
\label{table2}
\vspace{-1em}
\end{table}

\begin{figure}[t]
\centering
\includegraphics[width=0.99\linewidth]{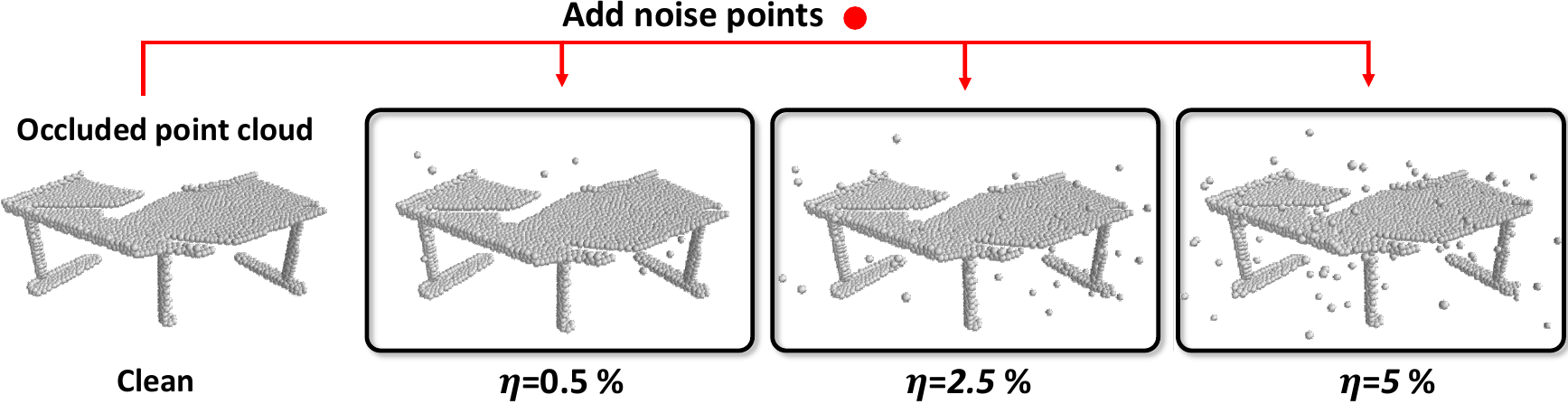}
\centering
\caption{Visualization of different ratios of noise points.}
\label{noisy_samples}
\vspace{-1.5em}
\end{figure}

The point cloud collected in the real scene will contain self-occlusion and unpredictable noise. 
To further ablate the noise tolerance of our PointMLS, we also establish varying noise levels to point clouds in ModelNet-O during testing~(see Fig.~\ref{noisy_samples}).
As presented in Tab.~\ref{table2}, while the accuracy of other methods decreases sharply as the ratio of noise points increases, our method maintains its accuracy with no more than a 10$\%$ drop. This phenomenon shows the remarkable robustness of PointMLS to noise.

\begin{table}[t]
\caption{Overall accuracy (OA) on ModelNet40~(N40) and ScanObjectNN~(SONN).
* denotes the re-implementation of the models that do not reach the reference results.
{[}P{]} denotes fine-tuning the models after self-supervised pre-training, which uses extra training data. Bold: best. Underline: second-best.}
\vspace{-0.5em}
\centering
\small
\setlength\tabcolsep{2.5mm}
\begin{tabular}{l|c|cc}
\toprule[0.15em]
Methods                                      & Venues     & MN40     & SONN      \\ \midrule[0.1em]
{[}P{]}Point-BERT~\cite{yu2022point-bert}    & CVPR'22    & 93.2     & 83.1      \\
{[}P{]}MaskPoint~\cite{liu2022masked}   & ECCV'22    & 93.8     & 84.3      \\
{[}P{]}Point-MAE~\cite{pang2022point-mae}    & ECCV'22    & 93.8     & 85.2      \\
{[}P{]}Point-M2AE~\cite{zhang2022point-m2ae} & NeurIPS'22 & 94.0     & 86.4      \\ 
{[}P{]}ACT~\cite{dong2022act}                & ICLR'23    & 93.7     & 88.2      \\
{[}P{]}I2P-MAE~\cite{zhang2023i2p-mae}       & CVPR'23    & 94.1     & 90.1      \\ \midrule
PointNet~\cite{qi2017pointnet}               & CVPR'17    & 89.2     & 63.4      \\
KCNet~\cite{shen2018kcnet}                   & CVPR'18    & 91.0     & -         \\
PAT~\cite{yang2019PAT}                       & CVPR'19    & 91.7     & -         \\
PointNet++~\cite{qi2017pointnet++}           & NeurIPS'17 & 91.9     & 75.4      \\
DGCNN~\cite{wang2019dynamic}                 & TOG'19     & 92.9     & 73.6      \\
PointCNN~\cite{li2018pointcnn}               & NeurIPS'18 & 92.5     & 75.1      \\
PointConv~\cite{wu2019pointconv}             & CVPR'19    & 92.5     & -         \\
PCT~\cite{guo2021pct}                        & CVM'21     & 92.6     & -         \\
PointASNL~\cite{yan2020pointasnl}            & CVPR'20    & 93.2     & -         \\
RS-CNN~\cite{liu2019rscnn}                   & CVPR'19    & 93.6     & -         \\
PointMLP*~\cite{ma2022pointmlp}              & ICLR'22    & 93.9     & 85.4      \\
RepSurf-U~\cite{ran2022surface}              & CVPR'22    & -        & 86.0      \\
CurveNet~\cite{xiang2021curvenet}            & ICCV'22    & \textbf{94.2} & -    \\
PointMeta~\cite{lin2023pointmetabase}        & CVPR'23    & -        & \textbf{87.9}   \\ \midrule
\textbf{PointMLS~(ours)}                     & -   & \underline{94.0}   & \underline{86.6} \\ 
\bottomrule[0.1em]
\end{tabular}
\centering
\label{table3}
\vspace{-1.5em}
\end{table}

\begin{table*}[t]
    \small
    \caption{Ablation study for the design choices of the proposed PointMLS.}
    \vspace{-0.5em}
    \centering
    \begin{minipage}{0.45\linewidth}
        \centering
        \subfloat[Ablation study on the sampling levels.]{
        \resizebox{1\textwidth}{!}{%
        \begin{tabular}{l|cccc|cc}
            \toprule[0.15em]
            \# & 1024 & 512 & 256 & 128 & mAcc~(\%) & OA~(\%)        \\ \midrule
            1  & \Checkmark    & -    & -    & -    & 76.7     & 77.7          \\
            2  & \Checkmark    & \Checkmark   & -    & -    & 77.3     & 78.2          \\
            3  & \Checkmark    & \Checkmark   & \Checkmark   & -    & 77.5     & 78.7          \\
            4  & \Checkmark    & \Checkmark   & \Checkmark   & \Checkmark  & \textbf{77.6}  & \textbf{78.9}  \\ \bottomrule[0.1em]
        \end{tabular}}}
    \end{minipage}
    \hspace{5mm}
    \begin{minipage}{0.5\linewidth}
        \renewcommand\arraystretch{1.05}
        \centering
        \subfloat[Ablation study on $MLP_{f}$.]{
        \resizebox{1\textwidth}{!}{%
        \begin{tabular}{l|l|cc}
        \toprule[0.15em]
            \# & $MLP_{f}$                     & mAcc~(\%) & OA~(\%) \\ \midrule
            1  & {[}128, 512{]}            & 76.3     & 76.9   \\
            2  & {[}64, 256, 512{]}         & 76.2     & 77.2   \\
            3  & {[}64, 128, 256, 512, 512{]} & \textbf{76.7}     & \textbf{77.7}   \\
            4  & {[}64, 128, 256, 512, 512, 512{]} & 76.5     & 77.6   \\ \bottomrule[0.1em]
        \end{tabular}}}
    \end{minipage}
    \vspace{3mm}

    \begin{minipage}{0.4\linewidth}
        \renewcommand\arraystretch{1}
        \subfloat[Sampling strategy.]{
            \resizebox{1\textwidth}{!}{%
            \begin{tabular}{l|cc|cc}
                \toprule[0.15em]
                \# & Before FA & Inside FA & mAcc~(\%) & OA~(\%) \\ \midrule
                1  & CPS      & CPS    & 84.3     & 85.6   \\
                2  & FPS      & CPS    & 84.5     & 86.0   \\
                3  & CPS      & FPS    & \textbf{85.0}     & \textbf{86.6}   \\ \bottomrule[0.1em]
            \end{tabular}}}
    \end{minipage}
    \hspace{2mm}
    \begin{minipage}{0.25\linewidth}
        \renewcommand\arraystretch{1.1}
        \subfloat[Annealing strategy of $\tau$.]{
        \resizebox{1\textwidth}{!}{%
            \begin{tabular}{c|cc}
                \toprule[0.15em]
                Scheduler & mAcc~(\%) & OA~(\%) \\ \midrule
                Lin       & 75.0        & 75.5      \\
                Exp       & 75.9        & 77.3      \\
                Cos       & \textbf{76.7}     & \textbf{77.7}  \\ \bottomrule[0.1em]
            \end{tabular}}}
    \end{minipage}
    \hspace{2mm}
    \begin{minipage}{0.3\linewidth}
        \renewcommand\arraystretch{1.25}
        \subfloat[Ablation study on the cost function.]{
        \resizebox{1\textwidth}{!}{%
            \begin{tabular}{cc|cc}
                \toprule[0.15em]
                $L_{cls}$ & $L_{spl}$ & mAcc~(\%) & OA~(\%) \\ \midrule
                \Checkmark    & -     & 76.1        & 77.2      \\
                \Checkmark    & \Checkmark    & \textbf{76.7}     & \textbf{77.7}   \\ \bottomrule[0.1em]
            \end{tabular}}}
    \end{minipage}
    \label{ablation_studies}
\vspace{-1em}
\end{table*}

\begin{table}[t]
    \caption{Effectiveness of the CPS module.}
    \small
    \centering
    \vspace{-0.5em}
    \setlength\tabcolsep{3mm}
    \begin{tabular}{l|l|ll}
        \toprule[0.15em]
        \#               & Combination                     & mAcc~(\%) & OA~(\%) \\ \midrule
        1 & FA + RS             & 76.1     & 76.5   \\
        2 & FA + FPS            & 76.3     & 76.7   \\
        3 & FA + CPS~(pre-trained)          & 76.6\blue{$^{\uparrow0.3}$}     & 77.4\blue{$^{\uparrow0.7}$}   \\
        4 & FA + CPS                                & \textbf{76.7\blue{$^{\uparrow0.4}$}}     & \textbf{77.7\blue{$^{\uparrow1.0}$}}   \\ \midrule
        5 & PointNet~\cite{qi2017pointnet} + FPS    & 65.6     & 66.8   \\
        6 & PointNet~\cite{qi2017pointnet} + CPS    & 68.3\blue{$^{\uparrow2.7}$}     & 70.0\blue{$^{\uparrow3.2}$}   \\ \midrule
        7 & DGCNN~\cite{wang2019dynamic} + FPS      & 74.7     & 75.9   \\
        8 & DGCNN~\cite{wang2019dynamic} + CPS      & 75.6\blue{$^{\uparrow0.9}$}     & 77.4\blue{$^{\uparrow1.5}$}   \\ \bottomrule[0.1em]
    \end{tabular}
\centering
\label{ablation_CPS}
\vspace{-1.5em}
\end{table}

\noindent
\textbf{Classification on Complete Point Clouds.}
To establish the generalizability of PointMLS to publicly available datasets that comprise complete point clouds, we also perform experiments on ModelNet40 and ScanObjectNN and achieve competitive performance~(94.0$\%$ and 86.6$\%$). 

\noindent
\textbf{ModelNet40.} 
For the testing phase, we adopt a voting strategy that uses random scaling and then averages the predictions, similar to what was done in~\cite{liu2019rscnn}.
As depicted in Tab.~\ref{table3}, our PointMLS achieves a respectable accuracy of 94.0$\%$. Although PointMLS does not achieve the highest performance, its results are non-negligible and competitive.
Note that although CurveNet has few network parameters and achieves state-of-art results, its inference cost is expensive~(can see in Supp.).
Experiments demonstrate that our PointMLS is versatile and can be applied not only to occluded point cloud datasets but also to complete point cloud datasets, with competitive performance.

\noindent
\textbf{ScanObjectNN.}
According to the results presented in Tab.~\ref{table3}, our PointMLS model achieves an impressive overall accuracy of 86.6$\%$, surpassing all other end-to-end 3D point cloud classification models.
Notably, the ScanObjectNN dataset used in our experiments is considered more challenging and realistic than the widely used ModelNet40 dataset.
Recently, several multi-stage methods modeled based on the BERT’s~\cite{devlin2018bert} training strategy were proposed and gained appealing results on ScanObjectNN.
However, these fine-tuned methods based on pre-trained models require additional data and more complicated training processes.
Despite this, our PointMLS model still outperforms most pre-trained models, falling only 1.6$\%$ short of the best-performing model, ACT~\cite{dong2022act}.

\subsection{Ablation Study and Analysis}

\noindent
\textbf{Effectiveness of the CPS Module.}
In this study, we assess the impact of the CPS module (Sec.~\ref{Section4.1}) on three baseline networks: FA module (Sec.~\ref{Section4.2}), PointNet~\cite{qi2017pointnet}, and DGCNN~\cite{wang2019dynamic}.
To ensure a fair comparison, we keep the number of input points consistent across all baselines.
Concretely, we sample 1024 points for the task networks using different sample strategies.
As presented in Tab.~\ref{ablation_CPS}, the FA module achieves OA scores of 76.5$\%$ and 76.7$\%$ using random and farthest point sampling methods~($\#$1,2), respectively.
We then replace the sampling strategy with our CPS module, resulting in an overall accuracy increase to 77.7$\%$~($\#$4).
Similarly, replacing the sampling module with CPS improves the overall accuracy of PointNet~\cite{qi2017pointnet} and DGCNN~\cite{wang2019dynamic} on the occluded point cloud dataset to varying degrees, as shown in Tab.~\ref{ablation_CPS} ($\#$5,6 and $\#$7,8).

We also examine the generalization ability of our CPS module. As shown in Tab.~\ref{ablation_CPS}~($\#$3), we adopt the CPS module trained on the complete point cloud dataset ModelNet40 as the sampler for the classification task on the occluded point cloud dataset ModelNet-C. The performance is better than the random and farthest point sampling methods, this result indicates that the CPS module possesses generalization ability during different datasets.

\noindent
\textbf{Different Combinations of Sampling Levels.}
To utilize both dense and sparse contextual information, we fuse multiple levels of sampled point clouds and report the results in Tab.~\ref{ablation_studies}(a).
Our analysis shows that the best classification results for occluded point clouds are obtained by fusing 4 levels of sub-point clouds. This suggests that each level of sparse point clouds contains unique information that complements the missing local information due to occlusion.

\begin{figure}[t]
\centering
\includegraphics[width=8.2cm]{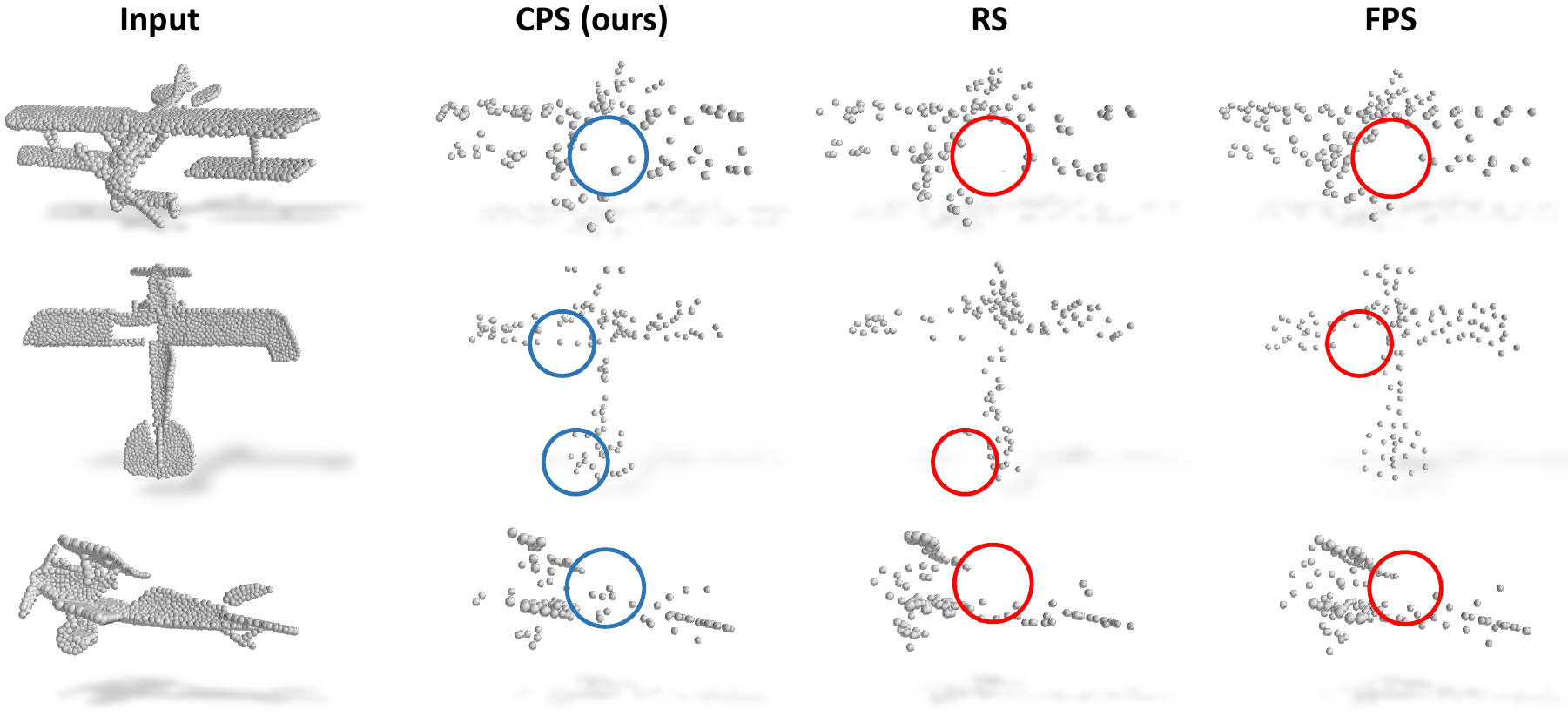}
\centering
\vspace{-0.5em}
\caption{Visualization of the sampling results of CPS, random sampling (RS), and farthest points sampling (FPS).
Circles of different colors represent different sampling effects. Blue is good, red is the opposite.}
\vspace{-0.5em}
\label{Occluded_pc}
\vspace{-1.5em}
\end{figure}

\noindent
\textbf{Dimension of $MLP_{f}$ in CPS module}. To determine the feature dimensions of $MLP_{f}$ belonging to the critical point sampling (CPS) module, we conducted several experiments. As presented in Tab.~\ref{ablation_studies}(b), the $MLP_{f}$ with dimensions [64, 128, 256, 512, 512] achieves the best mean class-average accuracy and overall accuracy. Note that this ablation study and the following don't utilize the multi-level manner.

\noindent
\textbf{Sampling Strategies in PointMLS.}
We conducted experiments on the ScanObjectNN to explore sampling strategies in PointMLS. As shown in Tab.~\ref{ablation_studies}(c), our original configuration ($\#$3) involves using CPS before the FA module, followed by the FPS strategy in the FA module's sampling process, resulting in an overall accuracy of 86.6$\%$. When we swap the CPS and FPS in our original configuration ($\#$2), we achieve an accuracy of 86.0$\%$. But this model is less robust than that of the original configuration, as its first batch of sampled points relies on FPS. Although CPS is theoretically superior to FPS, the Gumbel noises in the Gumbel-softmax of the CPS module serve as strong regularization, too much of which leads to an accuracy decrease~($\#$1).

\begin{figure}[t]
\centering
\includegraphics[width=7.8cm]{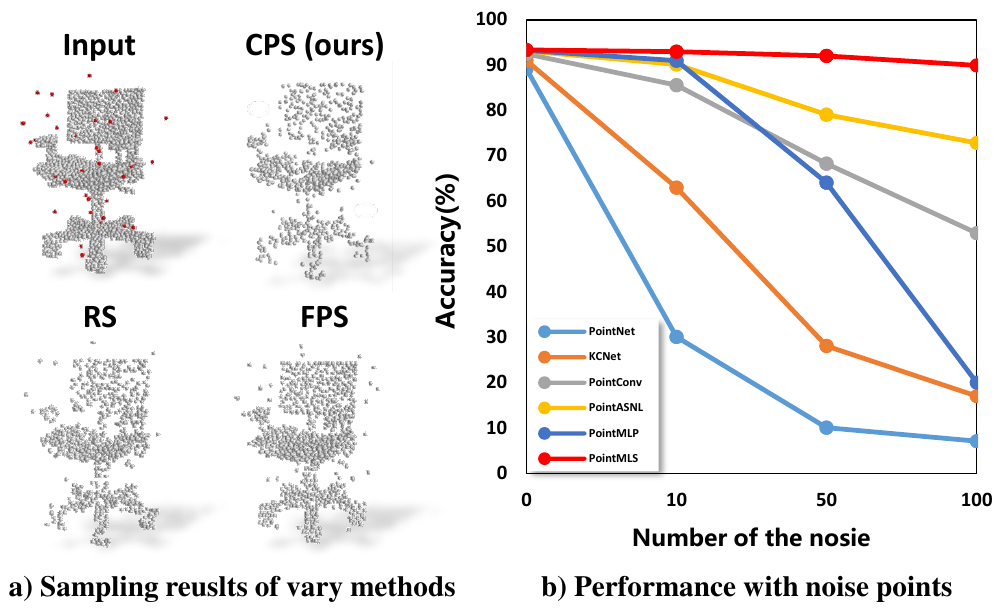}
\centering
\vspace{-0.5em}
\caption{a) Visualization of sampling results by using various methods on point clouds with noise points. b) Performance of different models on the ModelNet40 with noise points.}
\vspace{-0.5em}
\label{denoise}
\vspace{-1em}
\end{figure}

\noindent
\textbf{Temperature Profiles.}
We ablate the influence of several temperature parameter adjustment schemes, including cosine, linear, and exponential annealing. Our results in Ta.~\ref{ablation_studies}(d) demonstrate that cosine annealing achieves the best performance, allowing the CPS module to explore diverse sub-regions at the start of training and converge to the original point clouds at the end.

\noindent
\textbf{Objective Function.}
We analyze the impact of $L_{spl}$, a constraint that prevents the sampled point clouds from deviating too much from the original point clouds. Our results in Tab.~\ref{ablation_studies}(e) indicate that removing this constraint leads to a 0.5$\%$ decrease in the overall accuracy.

\noindent
\textbf{Visualization.}
Fig.~\ref{Occluded_pc} shows a visualization of the sampling results for CPS, RS, and FPS, where CPS preserves more structural information.
The circles indicate areas with occlusion. While sampling dense point clouds to sparse point clouds, CPS partially recovers these areas, whereas the other methods leave them blank. This demonstrates that CPS is able to better extract the overall structural feature of the occluded point clouds.

\subsection{Further Analysis.}
\noindent
\textbf{Robustness for Noisy Point Clouds.}
Despite the robustness testing on the occluded point cloud dataset ModelNet-O, we also conduct an experiment on the regular dataset ModelNet40.
Similar to~\cite{shen2018kcnet}, to further verify the robustness of our model by replacing the original cloud with normally distributed noise points during testing.
Note that our PointMLS does not apply the multi-level architecture here.
We also compare with PointNet~\cite{qi2017pointnet}, KC-Net~\cite{shen2018kcnet}, PointConv~\cite{wu2019pointconv}, PointMLP~\cite{ma2022pointmlp} and PointASNL~\cite{yan2020pointasnl}.
The experimental results are shown in Fig.~\ref{denoise}, which shows that our PointMLS decreases by less than 3$\%$ as the number of noise points increases to 100.
PointASNL~\cite{yan2020pointasnl}, which also includes a sampling module, is slightly inferior to our PointMLS because its AS module relies on FPS to sample the first batch of center points.
In contrast, Our model shows the strongest noise robustness due to the presence of the CPS module, which is insensitive to noise.

\begin{table}[]
\caption{Testing results on the proposed ModelNet-O dataset, after training on ModelNet40 with augmentation technologies.}
\centering
\small
\setlength\tabcolsep{1.5mm}
\vspace{-0.5em}
\begin{tabular}{l|c|cc}
\toprule[0.15em]
Methods                   & Augmentation & mAcc~($\%$) & OAc~($\%$)   \\ \midrule
\multirow{2}{*}{PointNet~\cite{qi2017pointnet}}  & Sample~\cite{hermosilla2018PDS}    & 10.2\red{$^{\downarrow55.4}$}    & 12.4\red{$^{\downarrow54.4}$}     \\
                                                 & Drop Local~\cite{ren2022ModelNet-C}   & 8.4\red{$^{\downarrow57.2}$}     & 10.1\red{$^{\downarrow56.7}$}     \\ \midrule
\multirow{2}{*}{DGCNN~\cite{wang2019dynamic}}    & Sample~\cite{hermosilla2018PDS}    & 10.8\red{$^{\downarrow63.9}$}    & 13.5\red{$^{\downarrow62.4}$}     \\
                                                 & Drop Local~\cite{ren2022ModelNet-C}   & 9.4\red{$^{\downarrow65.3}$}     & 11.1\red{$^{\downarrow64.8}$}     \\ \midrule
\multirow{2}{*}{PointMLP~\cite{ma2022pointmlp}}  & Sample~\cite{hermosilla2018PDS}    & 11.3\red{$^{\downarrow65.2}$}    & 12.5\red{$^{\downarrow64.4}$}     \\
                                                 & Drop Local~\cite{ren2022ModelNet-C}   & 9.2\red{$^{\downarrow67.0}$}     & 10.9\red{$^{\downarrow66.0}$}     \\ \bottomrule[0.1em]
\end{tabular}
\label{augmentation}
\vspace{-0.5em}
\end{table}

We also evaluate our PointMLS on robustness testing suit ModelNet-C~\cite{ren2022ModelNet-C} and present the results in the appendix. Our method achieves remarkable results.

\begin{figure}[t]
\centering
\includegraphics[width=0.99\linewidth]{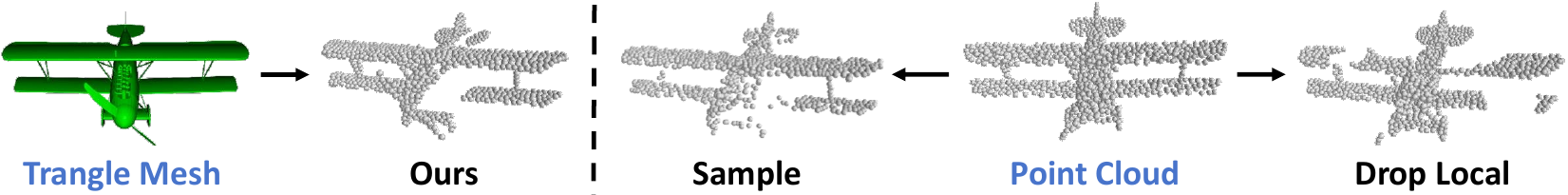}
\centering
\caption{Visualization of ModelNet-O and simple augmentations.}
\label{augmentation_samples}
\vspace{-1.5em}
\end{figure}

\noindent
\textbf{Compared to Simple Augmentations.} 
\cite{hermosilla2018PDS} introduces a simple data augmentation by orienting point sampling, while \cite{ren2022ModelNet-C} randomly drops local areas~(see Fig.~\ref{augmentation_samples}). Such two operations are similar to our ModelNet-O containing partially missing. However, our ModelNet-O refines occlusion detection through camera viewpoints and triangle mesh data, yielding more precise occluded point clouds. Furthermore, its cross-view assessment poses a challenging occlusion benchmark. To investigate the gap between ModelNet-O and simple augmentations, We evaluate methods~\cite{qi2017pointnet,wang2019dynamic,ma2022pointmlp}, trained on ModelNet40 with the above data augmentations, on the occluded benchmark ModelNet-O, as shown in Tab.~\ref{augmentation}. The significant results between ModelNet-O and simple augmentations highlight their distinctiveness, reinforcing the irreplaceability of our ModelNet-O.

\section{Conclusion}
In this paper, we propose a large-scale challenging occluded point cloud dataset ModelNet-O, which simulates the point cloud collecting process in real-world scenarios.
Then we introduce PointMLS, a robust method for classifying noisy point clouds under occlusion that employs multi-level sampling.
Our results demonstrate that PointMLS achieves state-of-the-art performance on the occluded point cloud dataset ModelNet-O and also shows excellent generalization ability on ModelNet40 and ScanObjectNN. 
In addition, PointMLS outperforms previous works in its ability to handle point clouds surrounded by noise points. 

\noindent
\textbf{Board Impact.} We propose a large-scale synthetic dataset ModelNet-O and an occlusion-aware point cloud classification model. We hope ModelNet-O can be a new benchmark for under-occlusion point cloud analysis and serve data-hungry methods.

{
    \small
    \bibliographystyle{ieeenat_fullname}
    \bibliography{main}
}
\clearpage
\setcounter{page}{1}
\maketitlesupplementary

\begin{figure}[t]
\centering
\includegraphics[width=8cm]{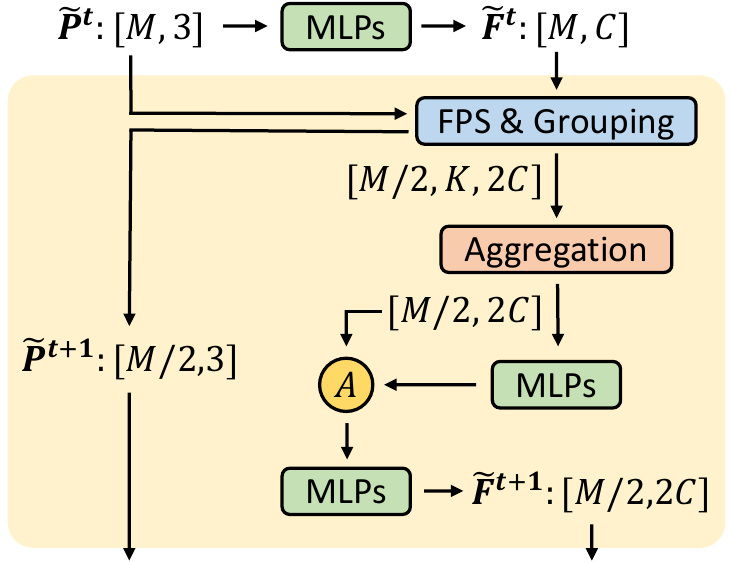}
\vspace{-1mm}
\caption{Illustration of Feature Aggregation(FA) Module. `A` means adding. }
\label{FA_architecture}
\vspace{-2mm}
\end{figure}

\section{Feature Aggregation (FA) Module Detail}
\label{sectionB}
We detail the architecture of the feature aggregation (FA) module in Fig.~\ref{FA_architecture}. In this module, the sampled points $\tilde{\mathbf{P}}^{t}$ and their corresponding features $\tilde{\mathbf{F}}^{t}$ are updated together using farthest point sampling (FPS) and the Maxpooling aggregation operations. Inspired by~\cite{qi2017pointnet,ma2022pointmlp}, we introduce the residual structure, which is then followed by the aggregation operation. We implement the aggregation function as a Maxpooling operation. During processing, the number of points gradually decreases by half, and the number of feature channels gradually doubles.

\section{More Robustness Analysis}
\label{sectionD}

\noindent
\textbf{Testing on ModelNet-C~\cite{ren2022ModelNet-C}.} Our study presents corruption error (CE) and relative corruption error (RCE) measures of 7 atomic corruptions based on the performance of DGCNN~\cite{wang2019dynamic} on ModelNet40, which is similar to this work~\cite{ren2022ModelNet-C}. We refer to the data from a previous study~\cite{ren2022ModelNet-C} and compare our PointMLS results to theirs. As indicated in Tab~\ref{mCE} and Tab.~\ref{RmCE}, PointMLS outperforms most of the ``Architectures" and ``Pre-training" methods. Note that the ``Augmentations" methods use complex data augmentation strategies during training, whereas our model simply adopts random scaling. Nonetheless, for the RCE metric, PointMLS achieves competitive results in the comparison with ``Augmentations" methods due to the remarkable robustness of the CPS module.

\begin{figure}[t]
\centering
\includegraphics[width=8cm]{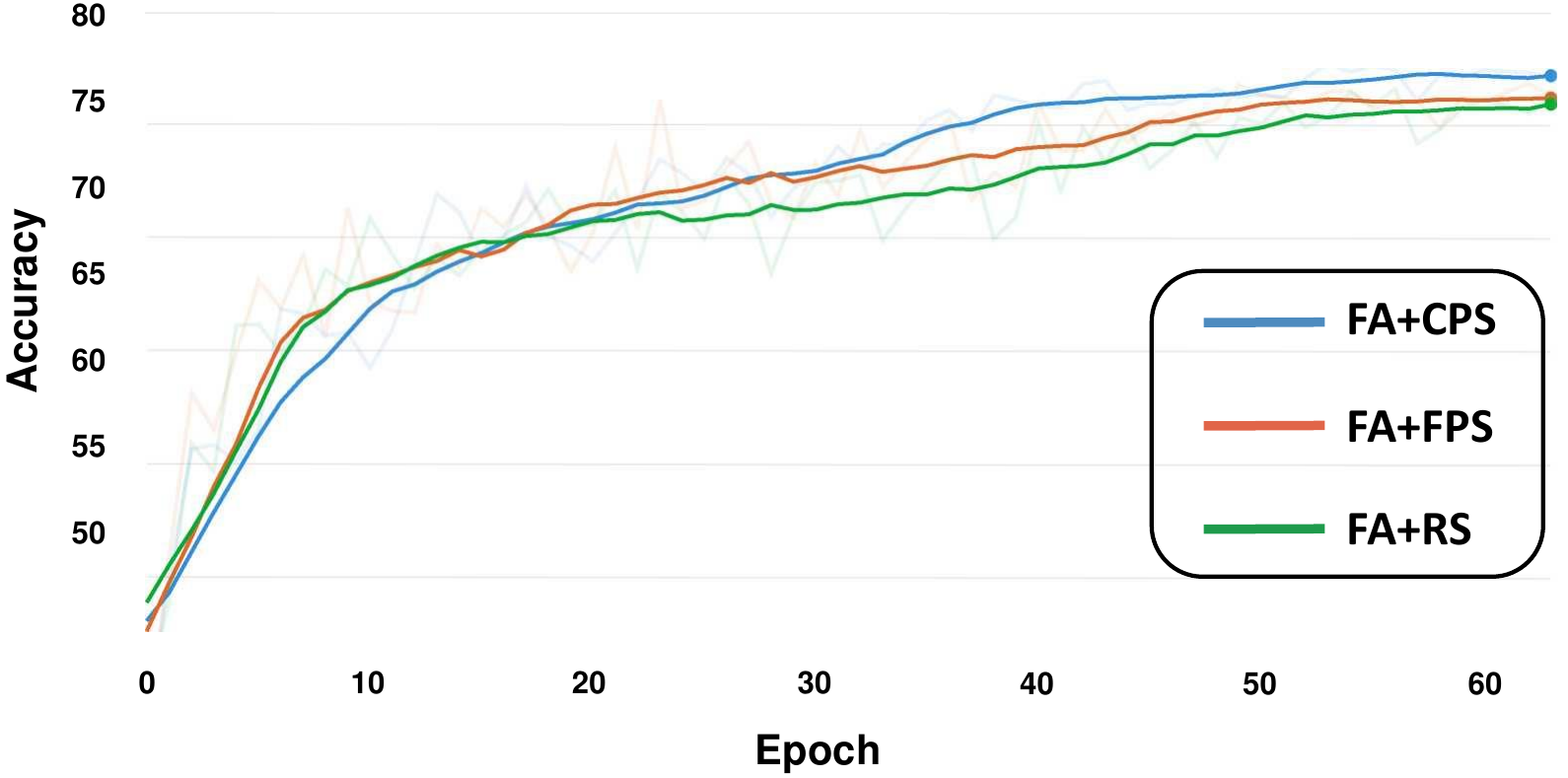}
\centering
\vspace{-0.5em}
\caption{The overall accuracy profiles of FA+CPS, FA+FPS and FA+RS. Our CPS converges faster.}
\label{Acc_profile}
\vspace{-1em}
\end{figure}

\noindent
\textbf{Visualization.} We present three overall accuracy trends in Fig.~\ref{Acc_profile} for three different sampling methods: FA+CPS, FA+FPS, and FA+RS, respectively. Our proposed CPS achieves the fastest convergence rate.

\begin{figure*}[t]
\centering
\includegraphics[width=0.99\textwidth]{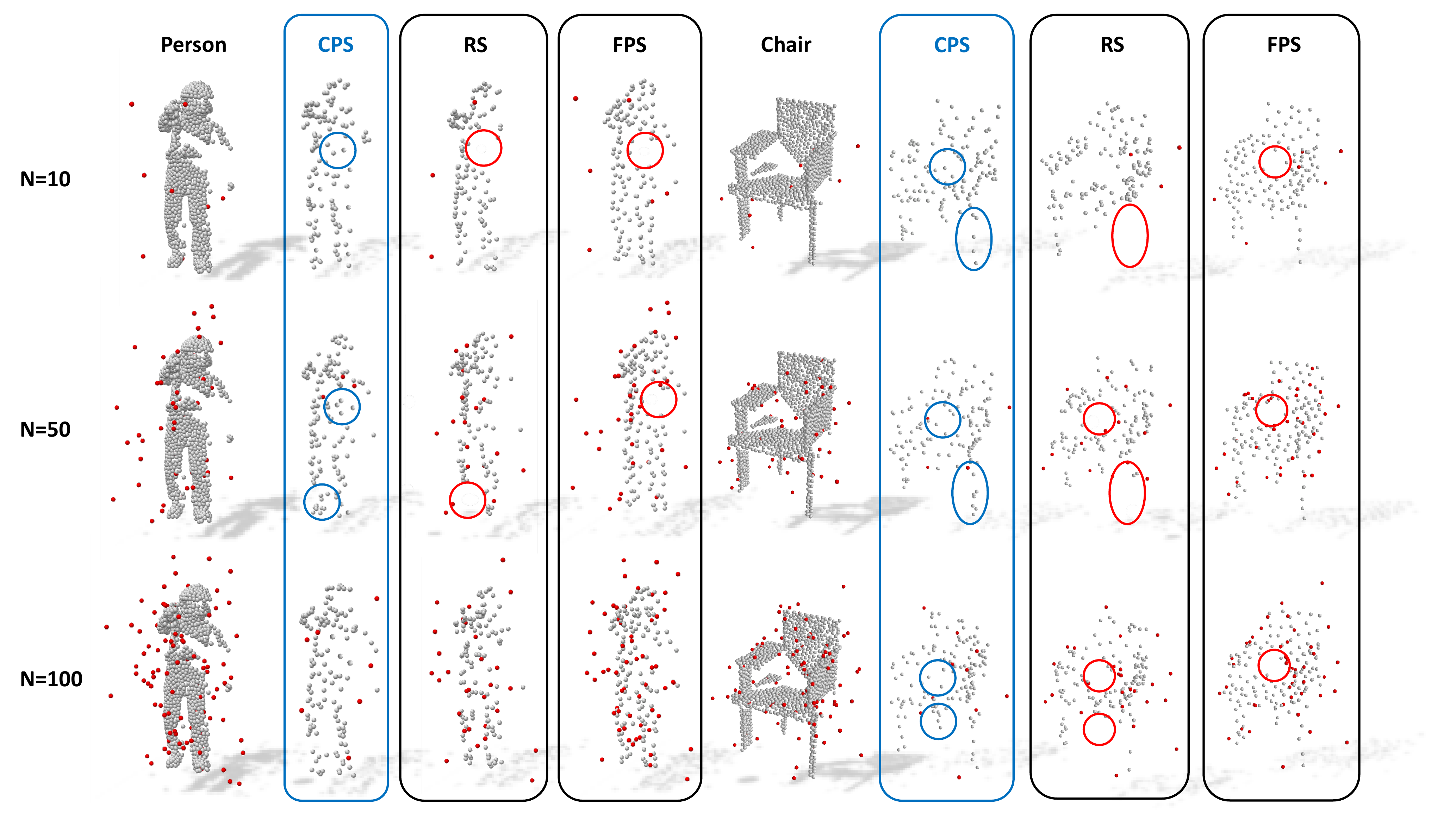}
\caption{Visualization of robustness testing on ModelNet-O. For this testing, We compare CPS, RS, and FPS sampling strategies. N denotes the number of noise points. We considered N to be 10, 50, and 100.}
\label{noise&occluded}
\end{figure*}

We compare the sampling outcomes of CPS, RS, and FPS visually under different noise levels. In order to see the sampled points more intuitively, we adopt hard sampling of Gumbel-softmax in the CPS module. In the case of little noise, our model can extract a sparse point cloud from the original dense point cloud while preserving its structural information. This approach can help mitigate the effects of local missing caused by self-occlusion. In contrast, the sampling processes of RS and FPS do not consider the structural information of the point cloud, but merely increase the sparsity of the point cloud. Additionally, during the sampling process, RS may overlook certain parts of the point cloud, as illustrated in the chair example of the second row in Fig.~\ref{noise&occluded}. In high-noise scenarios, compared to RS and FPS, our model can effectively isolate most of the noise and prevent the noise from affecting the subsequent point cloud classification tasks. In summary, our model exhibits strong robustness regardless of the amount of noise.

\section{Comparison of Running Speed and Size.}
Even with our multi-level sampling framework, our method outperforms CurveNet in running speed. We compare the size and running speed of the models mentioned in Tab.\ref{table2} of the main text, as shown in Tab.~\ref{comparison}. we report the speed of models by samples/second tested on one NVIDIA RTX 3080 Ti GPU.

\begin{table}[h]
\caption{comparison of model size and running speed.}
\centering
\tiny
\setlength\tabcolsep{0.6mm}
\renewcommand\arraystretch{1.3}
\begin{tabular}{c|ccccccccc}
\hline
      & PointNet & PointNet++ & DGCNN & PointConv & CurveNet & PCT & PointMLP & \begin{tabular}[c]{@{}c@{}}PointMLS\\ w/o ML\end{tabular} & \begin{tabular}[c]{@{}c@{}}PointMLS\\ w/ ML\end{tabular} \\ \hline
Train speed  & 375      & 412        & 406   & 126       & 98       & 185 & 122      & 105                                                       & 100                                                      \\
Test speed  & 411      & 531        & 616   & 181       & 150      & 246 & 312      & 252                                                       & 156                                                      \\
Params(M) & 1.6      & 1.5       & 1.8   & 19.6      & 2.1      & 2.9 & 13.2     & 11.2                                                      & 17.1                                                     \\ \hline
\end{tabular}
\vspace{-1em}
\label{comparison}
\end{table}

\section{More Analysis.}

\begin{figure}[t]
    \centering
    \begin{minipage}[!t]{0.48\linewidth}
        \subfloat[]{
        \includegraphics[width=4.1cm]{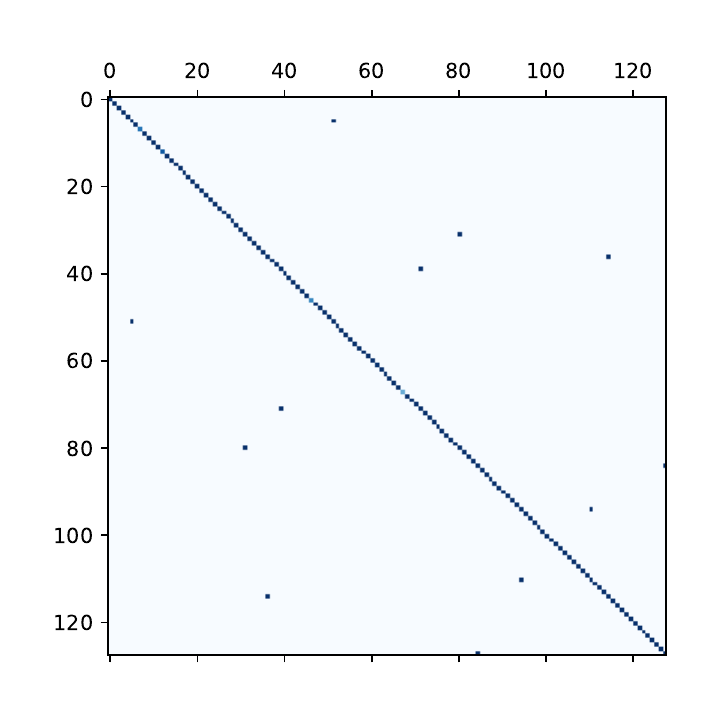}
        \centering
        \label{samplematrix}}
    \end{minipage}
    \begin{minipage}[!t]{0.48\linewidth}
        \subfloat[]{ 
        \includegraphics[width=4cm]{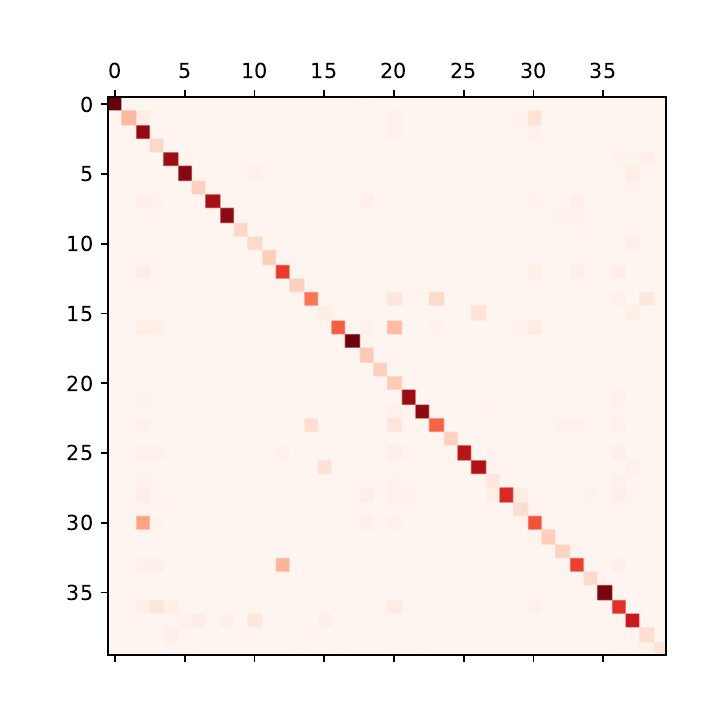}
        \centering 
        \label{each_acc_confusion_matrix}}
    \end{minipage}
    \vspace{-0.5em}
    \caption{Visualization of a) $WW^{T}$ matrix and b) the confusion matrix of results on ModelNet-O.}
    \vspace{-0.5em}
\label{fig7}
\vspace{-1em}
\end{figure}

\noindent
\textbf{Point-wise Weight Matrix.}
We visualize $WW^{T}$, where $W$ is the point-wise weight matrix with a sampling size of 128 points. As shown in Fig.\ref{fig7}(a), $\tilde{W}$ is nearly an identity matrix, meaning that $W$ can be treated as a permutation matrix. 
This ensures that most of the points will not be re-sampled, thus preserving the integrity of the original data and avoiding the collapse problem.

\noindent
\textbf{Error Analysis.}
We report per-category prediction results on the ModelNet-O testing set. Fig.\ref{fig7}(b) illustrates that our model can accurately classify most occluded point clouds, but some samples are misidentified into similar categories. This occurs because these categories become extremely similar when they are under occlusion. For instance, the "flower$\_$pot" and "plant" categories can appear indistinguishable when they are projected via one specific viewpoint and partial parts of them are occluded. This observation highlights the challenge of accurately classifying occluded point cloud datasets, such as ModelNet-O.

\begin{table*}[t]
\centering
\caption{Full results of corruption error (CE) on ModelNet40. Bold: best in the column of each strategy. Underline: second best in the column of ``architectures" methods.}
\vspace{-0.5em}
\setlength\tabcolsep{3.2mm}
\renewcommand\arraystretch{0.8}
\centering
\begin{tabular}{l|c|ccccccc}
\toprule[0.15em]
Method    & mCE$\downarrow$ & Scale  & Jitter & Drop-G  & Drop-L  & Add-G & Add-L & Rotate   \\ \midrule
\multicolumn{9}{c}{\textit{Architectures}}      \\ \midrule
DGCNN~\cite{wang2019dynamic}      & 1.000    & 1.000    & 1.000   & 1.000  & 1.000  & 1.000   & 1.000    & 1.000           \\
PointNet~\cite{qi2017pointnet}    & 1.422    & 1.266    & \textbf{0.642}   & \underline{0.500}  & 1.072  & 2.980   & 1.593    & 1.902           \\
PointNet2~\cite{qi2017pointnet++}    & 1.072    & 0.872    & 1.177   & 0.641  & 1.802  & \underline{0.614}   & 0.993    & 1.405           \\
RSCNN~\cite{liu2019rscnn}      & 1.130    & 1.074    & 1.171   & 0.806  & 1.517  & 0.712   & 1.153    & 1.479           \\
SimpleView~\cite{simpleview}    & 1.047    & 0.872    & 0.715   & 1.242  & 1.357  & 0.983   & \underline{0.844}    & 1.316           \\
GDANet~\cite{xu2021GDANet}      & 0.892    & \textbf{0.830}    & 0.839   & 0.794  & \underline{0.894}  & 0.871   & 1.036    & 0.981           \\
CurveNet~\cite{xiang2021curvenet}    & 0.927    & 0.872    & 0.725   & 0.710  & 1.024  & 1.346   & 1.000    & \textbf{0.809}           \\
PAConv~\cite{xu2021paconv}     & 1.104    & 0.904    & 1.465   & 1.000  & 1.005  & 1.085   & 1.298    & 0.967           \\
PCT~\cite{guo2021pct}        & 0.925    & 0.872    & 0.870   & 0.528  & 1.000  & 0.780   & 1.385    & 1.042           \\
RPC~\cite{ren2022ModelNet-C}        & \underline{0.863}    & \underline{0.840}    & 0.892   & \textbf{0.492}  & \textbf{0.797}  & 0.929   & 1.011    & 1.079           \\ \midrule
\textbf{PointMLS (ours)}  & \textbf{0.784} & 1.181 & \underline{0.652} & 0.565 & 0.913  & \textbf{0.522}  & \textbf{0.756}  & \underline{0.902}  \\ \midrule
\multicolumn{9}{c}{\textit{Self-supervised Pre-training}}            \\ \midrule
DGCNN~\cite{wang2019dynamic}+OcCo~\cite{wang2021occo}    & \textbf{1.047}    & 1.606    & \textbf{0.652}   & 0.903  & \textbf{1.039}  & \textbf{1.444}   & \textbf{0.847}     & \textbf{0.837}          \\
Point-BERT~\cite{yu2022point-bert}    & 1.248    & \textbf{0.936}    & 1.259   & \textbf{0.690}  & 1.150  & 1.932   & 1.440     & 1.326          \\ \midrule
\multicolumn{9}{c}{\textit{Augmentations}}       \\ \midrule
DGCNN~\cite{wang2019dynamic}+PointWOLF~\cite{kim2021pointWOLF}     & 0.814 & \textbf{0.926} & 0.864  & 0.988  & 0.874  & 0.807   & 0.764     & 0.479           \\
DGCNN~\cite{wang2019dynamic}+RSMix~\cite{lee2021RSMix}         & 0.745 & 1.319 & 0.873  & \textbf{0.653}  & 0.589  & \textbf{0.281}   & 0.629     & 0.870           \\
DGCNN~\cite{wang2019dynamic}+WOLFMix~\cite{ren2022ModelNet-C}    & \textbf{0.590} & 0.989    & 0.715  & 0.698  & \textbf{0.575}  & 0.285   & \textbf{0.415}     & \textbf{0.451}           \\
PointNet2~\cite{qi2017pointnet++}+PointMixUp~\cite{chen2020pointmixup}   & 1.028 & 1.670 & \textbf{0.712} & 0.802  & 1.812  & 0.458 & 0.615     & 1.130           \\ \bottomrule[0.1em]
\end{tabular}
\label{mCE}
\end{table*}

\begin{table*}[t]
\centering
\caption{Full results of relative corruption error (RCE) on ModelNet40. Bold: best in the column of each strategy. Underline: second best in the column of ``architectures" methods.}
\vspace{-0.5em}
\setlength\tabcolsep{3mm}
\renewcommand\arraystretch{0.8}
\centering
\begin{tabular}{l|c|ccccccc}
\toprule[0.15em]
Method               & RmCE$\downarrow$ & Scale & Jitter & Drop-G & Drop-L & Add-G & Add-L & Rotate \\ \midrule
\multicolumn{9}{c}{\textit{Architectures}}                                               \\ \midrule
DGCNN~\cite{wang2019dynamic}  & 1.000 & 1.000 & 1.000  & 1.000  & 1.000  & 1.000 & 1.000 & 1.000  \\
PointNet~\cite{qi2017pointnet}  & 1.488 & 1.300 & \underline{0.455}  & \textbf{0.178}  & 0.970  & 3.557 & 1.716 & 2.241  \\
PointNet2~\cite{qi2017pointnet++}  & 1.114 & 0.600 & 1.248  & 0.511  & 2.278  & \underline{0.502} & 1.010 & 1.645  \\
RSCNN~\cite{liu2019rscnn}    & 1.201 & 1.200 & 1.211  & 0.707  & 1.782  & 0.602 & 1.194 & 1.709  \\
SimpleView~\cite{simpleview} & 1.181 & 1.050 & 0.682  & 1.420  & 1.654  & 1.036 & \underline{0.851} & 1.574  \\
GDANet~\cite{xu2021GDANet}  & 0.865 & 0.600 & 0.822  & 0.753  & 0.895  & 0.864 & 1.090 & 1.028  \\
CurveNet~\cite{xiang2021curvenet} & 0.978 & 1.000 & 0.690  & 0.655  & 1.128  & 1.516 & 1.060 & \underline{0.794}  \\
PAConv~\cite{xu2021paconv}   & 1.211 & 1.050 & 1.649  & 1.057  & 1.083  & 1.158 & 1.458 & 1.021  \\
PCT~\cite{guo2021pct}   & 0.884 & 0.600 & 0.847  & 0.351  & 1.030  & 0.724 & 1.547 & 1.092  \\
RPC~\cite{ren2022ModelNet-C}  & \underline{0.778} & \underline{0.450} & 0.876  & 0.299  & \underline{0.714}  & 0.923 & 1.035 & 1.149  \\ \midrule
\textbf{PointMLS (ours)}  & \textbf{0.416} & \textbf{0.300} & \textbf{0.417}  & \underline{0.201}  & \textbf{0.632}  & \textbf{0.222} & \textbf{0.512} & \textbf{0.631}  \\ \midrule
\multicolumn{9}{c}{\textit{Self-supervised Pre-training}}                                                   \\ \midrule
DGCNN~\cite{wang2019dynamic}+OcCo~\cite{wang2021occo}  & 1.302 & 3.650 & \textbf{0.529}  & 0.839  & \textbf{1.030}  & \textbf{1.575} & \textbf{0.771} & \textbf{0.723}  \\
Point-BERT~\cite{yu2022point-bert}   & \textbf{1.262} & \textbf{0.500} & 1.322  & \textbf{0.534}  & 1.203  & 2.226 & 1.582 & 1.468  \\ \midrule
\multicolumn{9}{c}{\textit{Augmentations}}                                               \\ \midrule
DGCNN~\cite{wang2019dynamic}+PointWOLF~\cite{kim2021pointWOLF}  & 0.698 & \textbf{0.650} & 0.822  & 0.983  & 0.805  & 0.742 & 0.677 & 0.206  \\
DGCNN~\cite{wang2019dynamic}+RSMix~\cite{lee2021RSMix}  & 0.839 & 2.700 & 0.851  & 0.529  & 0.391  & \textbf{0.059} & 0.512 & 0.830  \\
DGCNN~\cite{wang2019dynamic}+WOLFMix~\cite{ren2022ModelNet-C}  & 0.485 & 1.250 & 0.653  & 0.603  & 0.383  & 0.072 & \textbf{0.229} & 0.206  \\
PointNet2~\cite{qi2017pointnet++}+PointMixUp~\cite{chen2020pointmixup} & 1.254 & 3.600 & \textbf{0.579}  & 0.655  & 2.180  & 0.226 & 0.418 & 1.121  \\
PCT~\cite{guo2021pct}+WOLFMix~\cite{ren2022ModelNet-C}   & 0.488 & 1.400 & 0.843  & \textbf{0.161}  & \textbf{0.271}  & 0.100 & 0.363 & 0.277  \\
GDANet~\cite{xu2021GDANet}+WOLFMix~\cite{ren2022ModelNet-C} & \textbf{0.439} & 0.950 & 0.880  & 0.379  & 0.361  & 0.109 & 0.239 & \textbf{0.156}  \\
RPC~\cite{ren2022ModelNet-C}+WOLFMix~\cite{ren2022ModelNet-C}  & 0.517 & 1.400 & 0.988  & 0.218  & 0.293  & 0.140 & 0.323 & 0.255  \\ \bottomrule[0.1em]
\end{tabular}
\label{RmCE}
\end{table*}

\end{document}